\documentclass{article}
\usepackage{spconf,amsmath,graphicx,hyperref}
\usepackage{amssymb}      

\usepackage{booktabs}     
\usepackage{multirow}     
\usepackage{graphicx}     
\usepackage{array}        

\usepackage[dvipsnames]{xcolor}

\newcommand{\pos}[1]{\textcolor{ForestGreen}{#1}}
\newcommand{\negv}[1]{\textcolor{red}{#1}}
\newcommand{\lightmidruleA}{\cmidrule[0.3pt](lr){1-12}}
\newcommand{\lightmidruleB}{\cmidrule[0.3pt](lr){1-14}}

\title{Does Joint-Embedding Predictive Architecture Pretraining Help\\ Time Series Forecasting?}
\name{Yutong Feng$^{1}$ \qquad Bowen Liao$^{2}$ \qquad See Kiong Ng$^{1}$ \qquad Yuxuan Liang$^{3*}$\thanks{* Corresponding author.}}
\address{$^{1}$National University of Singapore \quad
         $^{2}$South China University of Technology \\
         $^{3}$Hong Kong University of Science and Technology (Guangzhou)}
\begin{document}
%
\maketitle
\begin{abstract}
 Joint-embedding predictive architectures (JEPA) have emerged as a promising self-supervised pretraining paradigm for time series, learning representations by predicting target embeddings in latent space rather than reconstructing raw signals. Yet evidence on their benefits remains mixed, and most studies test only a single backbone or a narrow set of architectures, leaving unclear whether JEPA pretraining is a reliable improvement or one that depends heavily on the downstream model. We address this gap through a large scale evaluation of one JEPA instantiation across nine backbones and eleven benchmarks spanning temporal and spatio-temporal forecasting, the most extensive cross architecture assessment of JEPA for time series to date. We find that the benefit of this instantiation varies sharply across backbones, producing consistent gains for some architectures and consistent degradation for others, even on the same dataset. This pattern holds across both task families, indicating the variability is a general property of this instantiation rather than a dataset specific artifact worth accounting for when choosing a backbone in practice. 
\end{abstract}

\section{Introduction}
\label{sec:intro}

Self-supervised pretraining based on joint-embedding predictive architectures (JEPA) \cite{assran2023ijepa} offers a promising direction for time series representation learning. Existing self-supervised approaches for time series fall into two paradigms, each with a drawback: masked autoencoding methods \cite{nie2023patchtst,dong2023simmtm} reconstruct raw signal values directly, forcing the model to recover low-level detail that may be irrelevant to downstream tasks, while contrastive methods \cite{yue2022ts2vec,woo2022cost,zhang2022tfc} operate in representation space but depend on carefully designed data augmentations and positive/negative pairs that do not transfer cleanly across time series domains. JEPA avoids both limitations: it predicts target embeddings directly from context embeddings in latent space~\cite{lecun2022path}, without reconstruction or negative pairs, encouraging the encoder to capture predictable, semantically meaningful structure while discarding irrelevant low-level detail. Though originally developed for vision and language, this paradigm has since drawn growing interest in time series analysis~\cite{zhang2024self}, where recent work has begun applying it to both temporal forecasting and spatio-temporal learning \cite{li2024t,li2025hit}.

Reports on how much JEPA pretraining actually helps in these settings remain inconsistent, with some studies \cite{ennadir2025tsjepa,chaykowsky2025tf} showing clear gains and others \cite{potapov2026reconstruction,chemeris2026lenepa} showing negligible benefit or outright degradation. 
Most of this evidence comes from evaluations limited to a single backbone or a narrow family of architectures~\cite{he2026mts,verdenius2024lat,girgis2026time}, leaving open whether JEPA pretraining offers a dependable improvement or hinges on the choice of downstream model.
This gap is compounded by the unusually wide architectural spectrum of time series backbones, from convolutional and graph based encoders to attention based transformers~\cite{ahmed2023transformers,wen2022transformers}.

In this paper, we evaluate one JEPA instantiation (Section~\ref{sec:instantiation}) across a broad set of representative backbones and benchmarks spanning temporal and spatio-temporal forecasting, providing one of the most extensive cross architecture assessments of JEPA for time series to date.
We find that the benefit of this instantiation is highly inconsistent across backbones, substantially improving performance for some architectures while consistently degrading it for others, even on the same dataset and across both task families we study, indicating that this variability is not an artifact of a particular dataset or task type.
Our contributions include:

\begin{itemize}
    \item We conduct a large scale, cross architecture evaluation of JEPA pretraining on nine backbones and eleven datasets covering temporal and spatio-temporal forecasting, broadening prior evaluations that focus on one or two architectures.
    \item We show that the effect of JEPA pretraining on downstream performance varies sharply across backbones, ranging from consistent, sizable improvements to consistent degradation, and that this pattern is stable across both task families.
    \item We distill our findings into practical guidance on when JEPA pretraining helps and when practitioners should be cautious about applying it.
\end{itemize}


\clearpage

\section{Instantiating}
\label{sec:instantiation}

\begin{figure}
    \centering
    \includegraphics[width=0.9\linewidth]{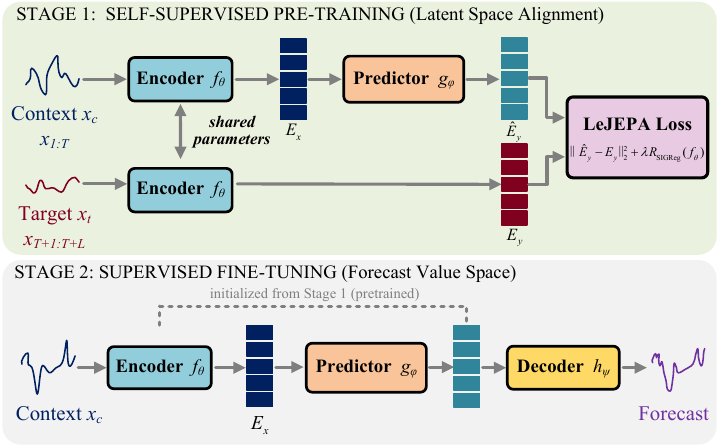}
    \caption{The two-stage JEPA instantiation for backbones. }
    \label{fig:jepa}
\end{figure}

Realizing the JEPA formalism introduced above in practice requires a specific architectural instantiation: a context view, a target view, an encoder, and a predictor. Existing forecasting backbones are architecturally diverse, spanning shallow linear models, deep attention based networks with explicit embedding modules, and graph based architectures that interleave spatial and temporal operations. If the effect of JEPA pretraining depends on backbone architecture, as we find later that it does, this effect must be measured under a single, consistent adaptation rather than under implementations that differ from one backbone to the next. To test whether JEPA is broadly effective for time series forecasting, rather than effective for one specifically designed architecture, we need a general and principled way to instantiate this formalism on top of pre-existing backbones, as shown in Figure \ref{fig:jepa}.


\noindent\textbf{Context and target views.}
Forecasting datasets already provide a natural, non-overlapping split between a historical window and a future window, so no additional masking scheme is explicitly required. We directly let the context view $x_c$ be the input sequence $x_{1:T}$ and the target view $x_t$ be the forecast horizon $x_{T+1:T+L}$, the ground truth label. This readily keeps JEPA pretraining compatible with existing forecasting dataloaders, since the pairs $(x_c, x_t)$ used for pretraining are the same input label pairs used for finetuning.

\noindent\textbf{Encoder.}
Given a backbone with encoder $f_\theta$, we obtain representations of both context and target views by applying the same encoder to $x_c$ and $x_t$:
\begin{equation}
    E_x = f_\theta(x_c), \qquad E_y = f_\theta(x_t).
\end{equation}
LeJEPA~\cite{balestriero2025lejepa} prevents collapse through the SIGReg regularizer rather than through stop gradients or a teacher student asymmetry, so a separate target encoder and a momentum update rule are unnecessary. 
A single shared encoder is sufficient for both branches, which removes the need to maintain and synchronize two encoder copies for every backbone family we consider. For most backbones, $f_\theta$ is not a module we add but the hidden representation the backbone already computes before its final output layer, such as an embedding and attention stack in a transformer based model or a stack of graph convolutions in a spatio-temporal model.

\noindent\textbf{Predictor and decoder.}
The predictor $g_\phi$ maps the context representation to a prediction of the target representation, $\hat{E}_y = g_\phi(E_x)$, which must match $E_y$ in shape so that a pretraining loss can be computed directly between them. In the general JEPA formulation~\cite{assran2023ijepa}, the predictor is also explicitly conditioned on relative position information between context and target. In the forecasting setting this extra conditioning is not needed, since the target window always follows the context window at a fixed, known offset, so we absorb it into a fixed, learned mapping from the context length to the target length. When a backbone already contains a module playing this role, such as a temporal projection operating in hidden space, we simply reuse it directly rather than adding a new one. The decoder $h_\psi$ maps the predicted target representation back to value space, $y_{\text{pred}} = h_\psi(\hat{E}_y)$, and corresponds to the backbone's existing output projection or regression head. It plays no role at all in the pretraining objective itself and is used only during finetuning, where it produces the forecast compared against the ground truth label.

\noindent\textbf{Decomposability criterion.}
Not every backbone admits a meaningful split into encoder, predictor, and decoder. We apply the instantiation above only when the encoder boundary can be identified without ambiguity, the shapes of $E_x$ and $E_y$ can be determined from the backbone's configuration alone, and the decoder boundary can be identified without ambiguity. When any of these conditions fails, for instance when a backbone maps directly from raw values to raw values through a single shallow transformation with no intermediate representation of distinct semantic content, we do not force a split. Introducing a layer purely to satisfy the encode predict decode template would align $\hat{E}_y$ and $E_y$ in an arbitrary space rather than a meaningful one, defeating the purpose of a representation space objective. Such backbones are excluded from backbone selection entirely, rather than retained within the study as supervised-only baselines.


\noindent\textbf{Training procedure.}
Pretraining and finetuning share the same computation pipeline: encode, then predict, and, only during finetuning, decode. During pretraining, the model receives $x_c$ and $x_t$ and is trained with the LeJEPA objective,
\begin{align}
    \mathcal{L}_{\text{pred}} &= \| \hat{E}_y - E_y \|_2^2 \\
    \mathcal{L}_{\text{LeJEPA}} &= \mathcal{L}_{\text{pred}} + \lambda \, \mathcal{R}_{\text{SIGReg}}(f_\theta)
\end{align}
computed entirely in representation space and without any negative pairs. During finetuning, the pretrained encoder and predictor are combined with the decoder and trained, or further trained, with a standard supervised forecasting loss computed between $y_{\text{pred}}$ and the ground truth label. This two-stage pipeline isolates the effect of JEPA pretraining from the choice of backbone, since the finetuning stage and the forecasting loss are kept identical to those of the fully supervised baseline for every backbone we consider.
\clearpage

\section{Experimental Setup}
\label{sec:setup}


\noindent\textbf{Datasets.}
We evaluate on eleven public benchmarks spanning two forecasting settings to ensure broad and representative coverage. For temporal forecasting, we use ETTh1, ETTh2, ETTm1, ETTm2, and Weather. 
For spatio-temporal forecasting, we use METR-LA, PEMS-BAY, PEMS03, PEMS04, PEMS07, and PEMS08. 
All datasets are split into training, validation, and test sets following standard practice for each benchmark. 
Each model receives a fixed input window of length $T$ and is trained to forecast a horizon of length $L$. 

\noindent\textbf{Backbones.}
We consider nine backbones spanning attention based and graph based architectures, with diverse architectural inductive biases.
All nine satisfy the decomposability criterion of Section~\ref{sec:instantiation} and are therefore pretrained with JEPA before finetuning. Architectures that fail this criterion, such as purely linear forecasters, were excluded during backbone selection rather than included as supervised-only baselines.

\noindent\textbf{Baselines and training protocol.}
For every JEPA compatible backbone, we compare two conditions. In the \textbf{w/o JEPA} condition, the backbone is trained from a random initialization directly on the supervised forecasting loss. In the \textbf{w/ JEPA} condition, the same backbone is first pretrained with the LeJEPA objective and then finetuned with the identical supervised forecasting loss. Total epochs are matched. The two conditions differ only in whether this pretraining stage is applied, since the finetuning stage, the loss function, and all downstream hyperparameters are kept identical between them. This isolates the contribution of JEPA pretraining from any other source of variation between the two conditions.


\noindent\textbf{Implementation details.}
For all backbones, the context length $T$ and forecast horizon $L$ 
are both set to 128, and the hidden representation dimension is 
fixed at 64. All models are optimized with Adam 
($\text{lr}=5\times10^{-4}$, weight decay $10^{-4}$).
We use effective batch size of 8, and clip gradient values to 5.
Both pretraining and finetuning are run for a maximum 
of 50 epochs with early stopping based on validation loss. The 
SIGReg regularization weight $\lambda$ is set to 0.1 and kept fixed 
across backbones, using 64 random slices.

\section{Experiment Results}
\label{sec:results}

We report results separately for the two task families introduced in Section~\ref{sec:setup}, following the protocol described there for a clean comparison. For each backbone, we compare the w/o JEPA and w/ JEPA conditions on MAE and RMSE across all datasets in the family, and summarize the direction and magnitude of the change as relative improvement.


\begin{table*}[t]
\centering
\caption{Time-series forecasting performance with and without JEPA pretraining. 
Results are reported as mean $\pm$ standard deviation over three random seeds. Lower values are better. Rel. Imp. is colored \pos{green} when JEPA improves performance and \negv{red} when it does not.
}
\label{tab:time_series_results}
\resizebox{\textwidth}{!}{%
\renewcommand{\arraystretch}{1.15}
\begin{tabular}{l c c c c c c c c c c c}
\toprule
\multirow{2}{*}{\textbf{Model}}
& \multirow{2}{*}{\textbf{Variant}}
& \multicolumn{2}{c}{\textbf{ETTh1}}
& \multicolumn{2}{c}{\textbf{ETTh2}}
& \multicolumn{2}{c}{\textbf{ETTm1}}
& \multicolumn{2}{c}{\textbf{ETTm2}}
& \multicolumn{2}{c}{\textbf{Weather}} \\
\cmidrule(lr){3-4}\cmidrule(lr){5-6}\cmidrule(lr){7-8}\cmidrule(lr){9-10}\cmidrule(lr){11-12}
& & \textbf{MAE} & \textbf{RMSE} & \textbf{MAE} & \textbf{RMSE} & \textbf{MAE} & \textbf{RMSE} & \textbf{MAE} & \textbf{RMSE} & \textbf{MAE} & \textbf{RMSE} \\
\midrule
\multirow{3}{*}{Crossformer}
& w/o JEPA & $1.56\pm0.01$ & $2.93\pm0.00$ & $2.84\pm0.02$ & $4.23\pm0.04$ & $1.26\pm0.01$ & $2.38\pm0.01$ & $2.11\pm0.01$ & $3.18\pm0.03$ & $11.03\pm0.05$ & $38.88\pm0.22$ \\
& w/ JEPA  & $1.53\pm0.01$ & $2.86\pm0.01$ & $2.79\pm0.00$ & $4.18\pm0.02$ & $1.25\pm0.00$ & $2.35\pm0.01$ & $2.10\pm0.00$ & $3.18\pm0.01$ & $10.57\pm0.02$ & $38.87\pm0.55$ \\
& Rel. Imp. & \pos{$1.71\%$} & \pos{$2.54\%$} & \pos{$1.55\%$} & \pos{$1.17\%$} & \pos{$1.34\%$} & \pos{$1.25\%$} & \pos{$0.36\%$} & \pos{$-0.00\%$} & \pos{$4.15\%$} & \pos{$0.01\%$} \\
\lightmidruleA
\multirow{3}{*}{iTransformer}
& w/o JEPA & $1.57\pm0.00$ & $2.96\pm0.01$ & $2.83\pm0.01$ & $4.24\pm0.00$ & $1.32\pm0.01$ & $2.40\pm0.00$ & $2.11\pm0.00$ & $3.22\pm0.00$ & $11.99\pm0.12$ & $41.76\pm0.06$ \\
& w/ JEPA  & $1.59\pm0.00$ & $2.98\pm0.01$ & $2.92\pm0.01$ & $4.36\pm0.00$ & $1.36\pm0.02$ & $2.44\pm0.03$ & $2.13\pm0.00$ & $3.22\pm0.00$ & $12.16\pm0.04$ & $41.06\pm0.30$ \\
& Rel. Imp. & \negv{$-1.52\%$} & \negv{$-0.70\%$} & \negv{$-2.93\%$} & \negv{$-2.72\%$} & \negv{$-2.73\%$} & \negv{$-1.67\%$} & \negv{$-0.68\%$} & \negv{$-0.12\%$} & \negv{$-1.40\%$} & \pos{$1.66\%$} \\
\lightmidruleA
\multirow{3}{*}{Autoformer}
& w/o JEPA & $1.80\pm0.05$ & $3.23\pm0.06$ & $3.39\pm0.05$ & $4.88\pm0.09$ & $1.57\pm0.01$ & $2.73\pm0.04$ & $2.35\pm0.01$ & $3.44\pm0.00$ & $21.30\pm0.95$ & $59.62\pm1.73$ \\
& w/ JEPA  & $1.70\pm0.08$ & $3.11\pm0.13$ & $3.48\pm0.17$ & $4.94\pm0.19$ & $1.50\pm0.00$ & $2.63\pm0.00$ & $2.35\pm0.01$ & $3.43\pm0.01$ & $24.61\pm1.62$ & $66.78\pm4.25$ \\
& Rel. Imp. & \pos{$5.41\%$} & \pos{$3.73\%$} & \negv{$-2.66\%$} & \negv{$-1.19\%$} & \pos{$4.66\%$} & \pos{$3.70\%$} & \negv{$-0.08\%$} & \pos{$0.16\%$} & \negv{$-15.54\%$} & \negv{$-12.00\%$} \\
\lightmidruleA
\multirow{3}{*}{FEDformer}
& w/o JEPA & $1.64\pm0.01$ & $2.99\pm0.02$ & $3.11\pm0.01$ & $4.52\pm0.03$ & $1.46\pm0.02$ & $2.53\pm0.01$ & $2.38\pm0.00$ & $3.48\pm0.00$ & $18.07\pm0.55$ & $52.53\pm2.19$ \\
& w/ JEPA  & $1.62\pm0.00$ & $2.97\pm0.00$ & $3.06\pm0.04$ & $4.44\pm0.05$ & $1.44\pm0.03$ & $2.51\pm0.03$ & $2.33\pm0.04$ & $3.38\pm0.06$ & $16.92\pm0.87$ & $48.39\pm1.98$ \\
& Rel. Imp. & \pos{$0.76\%$} & \pos{$0.80\%$} & \pos{$1.68\%$} & \pos{$1.70\%$} & \pos{$0.91\%$} & \pos{$0.93\%$} & \pos{$2.40\%$} & \pos{$2.64\%$} & \pos{$6.37\%$} & \pos{$7.89\%$} \\
\lightmidruleA
\multirow{3}{*}{PatchTST}
& w/o JEPA & $1.78\pm0.04$ & $3.39\pm0.08$ & $2.88\pm0.01$ & $4.32\pm0.04$ & $1.38\pm0.04$ & $2.54\pm0.06$ & $2.23\pm0.01$ & $3.37\pm0.03$ & $11.53\pm0.07$ & $38.36\pm0.43$ \\
& w/ JEPA  & $1.68\pm0.04$ & $3.20\pm0.09$ & $2.91\pm0.03$ & $4.32\pm0.03$ & $1.40\pm0.01$ & $2.54\pm0.00$ & $2.26\pm0.02$ & $3.45\pm0.02$ & $10.93\pm0.08$ & $37.66\pm0.03$ \\
& Rel. Imp. & \pos{$5.61\%$} & \pos{$5.58\%$} & \negv{$-1.02\%$} & \negv{$-0.07\%$} & \negv{$-1.47\%$} & \negv{$-0.03\%$} & \negv{$-1.43\%$} & \negv{$-2.33\%$} & \pos{$5.22\%$} & \pos{$1.83\%$} \\
\bottomrule
\end{tabular}%
}
\end{table*}

\begin{table*}[t]
\centering
\vspace{-7pt}
\caption{Spatio-temporal forecasting performance with and without JEPA pretraining. 
}
\label{tab:spatio_temporal_results}
\resizebox{\textwidth}{!}{%
\renewcommand{\arraystretch}{1.15}
\begin{tabular}{l c c c c c c c c c c c c c}
\toprule
\multirow{2}{*}{\textbf{Model}}
& \multirow{2}{*}{\textbf{Variant}}
& \multicolumn{2}{c}{\textbf{METR-LA}}
& \multicolumn{2}{c}{\textbf{PEMS-BAY}}
& \multicolumn{2}{c}{\textbf{PEMS03}}
& \multicolumn{2}{c}{\textbf{PEMS04}}
& \multicolumn{2}{c}{\textbf{PEMS07}}
& \multicolumn{2}{c}{\textbf{PEMS08}} \\
\cmidrule(lr){3-4}\cmidrule(lr){5-6}\cmidrule(lr){7-8}\cmidrule(lr){9-10}\cmidrule(lr){11-12}\cmidrule(lr){13-14}
& & \textbf{MAE} & \textbf{RMSE} & \textbf{MAE} & \textbf{RMSE} & \textbf{MAE} & \textbf{RMSE} & \textbf{MAE} & \textbf{RMSE} & \textbf{MAE} & \textbf{RMSE} & \textbf{MAE} & \textbf{RMSE} \\
\midrule
\multirow{3}{*}{ASTGCN}
& w/o JEPA & $5.55\pm0.17$ & $9.29\pm0.20$ & $2.60\pm0.01$ & $5.28\pm0.04$ & $22.99\pm0.14$ & $37.05\pm0.15$ & $27.41\pm0.66$ & $41.56\pm0.81$ & $32.43\pm0.01$ & $50.86\pm0.19$ & $21.54\pm0.33$ & $34.21\pm0.24$ \\
& w/ JEPA  & $5.84\pm0.09$ & $9.92\pm0.16$ & $2.68\pm0.04$ & $5.24\pm0.07$ & $23.95\pm0.29$ & $38.16\pm0.08$ & $29.44\pm0.20$ & $43.99\pm0.38$ & $33.56\pm0.21$ & $52.21\pm0.35$ & $23.77\pm0.36$ & $36.86\pm0.62$ \\
& Rel. Imp. & \negv{$-5.26\%$} & \negv{$-6.80\%$} & \negv{$-3.12\%$} & \pos{$0.68\%$} & \negv{$-4.16\%$} & \negv{$-3.00\%$} & \negv{$-7.41\%$} & \negv{$-5.85\%$} & \negv{$-3.48\%$} & \negv{$-2.65\%$} & \negv{$-10.36\%$} & \negv{$-7.77\%$} \\
\lightmidruleB
\multirow{3}{*}{STAEformer}
& w/o JEPA & $4.66\pm0.13$ & $8.67\pm0.01$ & $2.41\pm0.02$ & $5.02\pm0.05$ & $24.56\pm0.04$ & $37.93\pm0.08$ & $26.51\pm0.09$ & $39.88\pm0.10$ & $30.77\pm0.18$ & $47.75\pm0.66$ & $21.00\pm0.07$ & $33.54\pm0.14$ \\
& w/ JEPA  & $4.81\pm0.13$ & $9.21\pm0.10$ & $2.54\pm0.03$ & $5.23\pm0.07$ & $25.90\pm1.21$ & $40.03\pm1.36$ & $27.67\pm0.03$ & $41.37\pm0.07$ & $30.46\pm3.23$ & $47.54\pm3.02$ & $21.32\pm0.80$ & $33.61\pm1.11$ \\
& Rel. Imp. & \negv{$-3.24\%$} & \negv{$-6.17\%$} & \negv{$-5.35\%$} & \negv{$-4.18\%$} & \negv{$-5.45\%$} & \negv{$-5.54\%$} & \negv{$-4.38\%$} & \negv{$-3.74\%$} & \pos{$1.00\%$} & \pos{$0.43\%$} & \negv{$-1.53\%$} & \negv{$-0.20\%$} \\
\lightmidruleB
\multirow{3}{*}{STTN}
& w/o JEPA & $5.62\pm0.05$ & $9.92\pm0.06$ & $3.09\pm0.03$ & $6.04\pm0.15$ & $32.31\pm0.42$ & $47.60\pm0.27$ & $37.02\pm1.70$ & $51.66\pm1.59$ & $51.21\pm1.20$ & $72.48\pm2.31$ & $29.48\pm0.26$ & $43.97\pm0.31$ \\
& w/ JEPA  & $5.91\pm0.20$ & $9.93\pm0.07$ & $3.08\pm0.13$ & $5.84\pm0.01$ & $30.94\pm0.30$ & $46.46\pm0.92$ & $34.95\pm1.80$ & $49.69\pm2.31$ & $44.90\pm1.83$ & $63.29\pm3.20$ & $30.84\pm0.26$ & $44.64\pm0.22$ \\
& Rel. Imp. & \negv{$-5.34\%$} & \negv{$-0.02\%$} & \pos{$0.31\%$} & \pos{$3.43\%$} & \pos{$4.25\%$} & \pos{$2.39\%$} & \pos{$5.59\%$} & \pos{$3.81\%$} & \pos{$12.31\%$} & \pos{$12.67\%$} & \negv{$-4.62\%$} & \negv{$-1.54\%$} \\
\lightmidruleB
\multirow{3}{*}{STGCN}
& w/o JEPA & $5.12\pm0.07$ & $9.40\pm0.18$ & $2.58\pm0.05$ & $5.40\pm0.07$ & $23.54\pm0.13$ & $37.27\pm0.02$ & $24.53\pm0.02$ & $38.48\pm0.04$ & $27.53\pm0.12$ & $45.35\pm0.01$ & $20.14\pm0.15$ & $33.30\pm0.42$ \\
& w/ JEPA  & $4.71\pm0.02$ & $8.79\pm0.01$ & $2.36\pm0.01$ & $5.01\pm0.00$ & $22.61\pm0.12$ & $35.92\pm0.09$ & $24.55\pm0.00$ & $38.41\pm0.17$ & $26.63\pm0.09$ & $44.32\pm0.07$ & $20.04\pm0.08$ & $32.27\pm0.26$ \\
& Rel. Imp. & \pos{$8.02\%$} & \pos{$6.44\%$} & \pos{$8.26\%$} & \pos{$7.22\%$} & \pos{$3.93\%$} & \pos{$3.62\%$} & \negv{$-0.07\%$} & \pos{$0.19\%$} & \pos{$3.26\%$} & \pos{$2.28\%$} & \pos{$0.49\%$} & \pos{$3.10\%$} \\
\bottomrule
\end{tabular}%
}
\end{table*}

\noindent\textbf{Temporal Forecasting.}
Table~\ref{tab:time_series_results} reports results on the five temporal forecasting benchmarks. The effect of JEPA pretraining is not uniform across backbones, showing backbone dependence. Crossformer and FEDformer show a positive relative improvement on both metrics for every dataset, with FEDformer reaching a gain of 6.37\% in MAE and 7.89\% in RMSE on Weather. iTransformer shows the opposite pattern, with a negative relative improvement on both metrics for four of the five datasets, degrading by as much as 2.93\% in MAE on ETTh2. Autoformer and PatchTST fall between these two extremes, improving on some datasets while degrading on others, including a degradation of 15.54\% in MAE and 12.00\% in RMSE for Autoformer on Weather, the single largest negative effect observed in this task family.

\noindent\textbf{Spatio-Temporal Forecasting.}
Table~\ref{tab:spatio_temporal_results} reports results on the six spatio-temporal forecasting benchmarks. The spread between backbones is substantially wider here than in the temporal setting. STGCN shows a consistent positive pattern, improving on both metrics for five of six datasets, with gains reaching 8.26\% in MAE and 7.22\% in RMSE on PEMS-BAY. In contrast, ASTGCN and STAEformer degrade on both metrics for the majority of datasets, with ASTGCN losing 10.36\% in MAE and 7.77\% in RMSE on PEMS08. STTN shows a mixed pattern, improving on four of six datasets while degrading on the remaining two.



\section{Analysis \& Discussion}
\label{sec:analysis}


\noindent\textbf{A pattern related to architecture type.}
Backbones built primarily around dense, global self-attention, including iTransformer, STAEformer, and the attention augmented ASTGCN, show negative or negligible relative improvement on most datasets in their task family. Backbones whose mixing is instead structurally constrained, by graph or convolutional structure in STGCN, by frequency domain decomposition in FEDformer, or by segment and dimension restricted attention in Crossformer, show positive relative improvement on most or all datasets in their family. The split is not perfect, since Autoformer, PatchTST, and STTN combine attention with other components and show a mixed pattern. This split appears independently in both temporal and spatio-temporal forecasting, despite different backbones, inputs, and datasets in each family, which would be a coincidence if the two settings were unrelated, suggesting recurring patterns. We report this as an observation rather than an explanation, since confirming it would require experiments.

\noindent\textbf{Practical implications.}
Taken across all backbones and datasets in both task families, the gap between the largest positive and largest negative relative improvement we observe exceeds 25 percentage points, a swing large enough to matter in practice rather than a narrow band of noise around zero. For example, STGCN gains consistently while Autoformer loses over 15\% on Weather, reversing the preferred model. Applying JEPA pretraining without regard to backbone choice can therefore change which model appears best on a given benchmark, and can hide a strong backbone behind a degraded result as easily as it can reveal one. Practitioners should treat JEPA pretraining as backbone dependent rather than universally beneficial, and verify its effect on their own architecture rather than assume it transfers from another.

These observations return us to the question posed in this paper's title: JEPA does not work uniformly for time series self-supervised learning. Whether it works depends on the backbone it is paired with, and the difference between a favorable and unfavorable pairing is large enough to be decisive in practice, without a full account of why that difference arises. One possible explanation is that backbones differ in capturing predictive structure shared across context and target windows, making some representations better suited to JEPA pretraining under otherwise identical training conditions. We view this as a hypothesis motivating targeted ablations rather than a confirmed mechanism that future work should test directly.



\section{Conclusion \& Future Works}
\label{sec:conclusion}

We set out to answer whether joint-embedding predictive architectures are effective for time series self-supervised learning. Across nine backbones and eleven datasets spanning temporal and spatio-temporal forecasting, we find that the answer depends heavily on the backbone being pretrained. JEPA pretraining delivers consistent, sizable gains for some architectures and consistent degradation for others, with a gap large enough to change which model would be selected in practice. This inconsistency held across both task families, indicating that it reflects a property of the pretraining method rather than an artifact of any single dataset or task type. We did not aim to identify the mechanism behind this variability, only to establish that it exists and matters.

We observed a pattern linking this variability to architecture type. Backbones with dense, global self-attention benefit less than those with structurally constrained mixing, though the split is not perfect. Future work should test this pattern directly by varying inductive bias within a single backbone family. A natural next step would be to ablate the predictor and the SIGReg regularizer independently, to determine whether the variability originates in the pretraining objective itself or in how cleanly each backbone exposes an encoder and decoder boundary. Until these questions are answered, JEPA pretraining for time series should be adopted with the choice of backbone in mind, not assumed to help by default.

\bibliographystyle{IEEEbib}
\bibliography{bib/references}

\end{document}